\documentclass[runningheads]{llncs}

\usepackage[utf8]{inputenc}

\usepackage{cite}
\usepackage{algorithmic}
\usepackage{textcomp}

\usepackage{amssymb}

\usepackage{amsfonts} 

\usepackage{paralist}

\usepackage{blindtext}
\usepackage{mathtools}
\usepackage{booktabs} 
\usepackage{graphicx}
\usepackage{subcaption}
\usepackage{hyperref}
\usepackage{multicol}
\usepackage{longtable}
\usepackage{tabto}
\usepackage{comment}
\usepackage{xspace}
\usepackage{hyperref}
\usepackage{framed}
\usepackage{tcolorbox}
\usepackage{float}

\usepackage{blindtext}
\usepackage[ruled,vlined]{algorithm2e}

\SetAlCapSkip{1em}

\SetKwInput{KwInput}{Input}
\SetKwInput{KwOutput}{Output}

\makeatletter
\newcommand*{\encircle}[1]{\relax\ifmmode\mathpalette\@encircle@math{#1}\else\@encircle{#1}\fi}
\newcommand*{\@encircle@math}[2]{\@encircle{$\m@th#1#2$}}
\newcommand*{\@encircle}[1]{%
  \tikz[baseline,anchor=base]{\node[draw,circle,outer sep=0pt,inner sep=.2ex] {#1};}}
\makeatother

\usepackage{array}

\usepackage[thinlines]{easytable}

\usepackage{tikz}
\usepackage{collcell}
\usepackage{rotating}

\usepackage[]{quoting}

\usepackage{enumitem}

\usepackage{xcolor}
\hypersetup{
    colorlinks,
    linkcolor={black!50!black},
    urlcolor={blue!80!black}
}
\usepackage{ifthen}
\newboolean{showcomments}
\setboolean{showcomments}{false} 
\ifthenelse{\boolean{showcomments}}
  {\newcommand{\nb}[2]{
    \fcolorbox{red}{yellow}{\bfseries\sffamily\scriptsize#1}
    {\sf\small$\blacktriangleright$\textit{#2}$\blacktriangleleft$}
  }
  
  }
  {\newcommand{\nb}[2]{}
  
  }
\newcommand\pierg[1]{\nb{PG}{\textcolor{red}{#1}}}

\usepackage[normalem]{ulem} 

\newcommand{\ins}[1]{\textcolor{black}{#1}} 

\makeatletter
\newcommand*{\balancecolsandclearpage}{%
  \close@column@grid
  \clearpage
  \twocolumngrid
}
\makeatother

\usepackage{amsmath,scalerel}

\makeatletter
\newcommand*\bigcdot{\mathpalette\bigcdot@{.5}}
\newcommand*\bigcdot@[2]{\mathbin{\vcenter{\hbox{\scalebox{#2}{$\m@th#1\bullet$}}}}}
\makeatother

\newcommand{\NAME}{CR$^3$\xspace}

\begin{document}


\title{Contract-Based Specification Refinement and Repair for Mission Planning}


\author{%
Piergiuseppe Mallozzi\inst{1}  \and 
Inigo Incer\inst{1}
Pierluigi Nuzzo\inst{2} \and 
Alberto Sangiovanni-Vincentelli\inst{1}
}%
\institute{
UC Berkeley, USA\\
\email{mallozzi@berkeley.edu}, \email{inigo@berkeley.edu}, \email{nuzzo@usc.edu},  \email{alberto@berkeley.edu}
\and
University of Southern California, USA}








\maketitle


\begin{abstract}
We address the problem of modeling, refining, and repairing formal specifications for robotic missions using assume-guarantee contracts.
We show how to model mission specifications at various levels of abstraction and implement them using a library of pre-implemented specifications.
Suppose the specification cannot be met using components from the library. In that case, we compute a proxy for the best approximation to the specification that can be generated using elements from the library. Afterward, we propose a systematic way to either 1) \textit{search} for and refine the `missing part' of the specification that the library cannot meet or 2) \textit{repair} the current specification such that the existing library can refine it. Our methodology for searching and repairing mission requirements leverages the quotient, separation, composition, and merging operations between contracts. 
\end{abstract}

\section{Introduction}


Mission specification is a formulation of the mission in a formal (logical) language with precise semantics~\cite{specificationpatternstse2015}. Many results in the literature highlight the advantages of specifying robotic missions in a temporal logic language, like linear temporal logic (LTL) or computation tree logic (CTL)
~\cite{maozsynthesis,doi:10.1177/0278364914546174,finucane2010ltlmop,menghi2018multi,fainekos2009temporal,DBLP:journals/corr/MaozR16,MaozFSE,Shoukry17}.
Producing suitable \textit{implementations} of mission specifications is the problem of finding a policy to be followed by the robot such that the mission specification is always satisfied.
For specifications in LTL, reactive synthesis can automatically generate correct-by-construction implementations from a given specification~\cite{kress2009temporal,maniatopoulos2016reactive,finucane2010ltlmop, 9551578, 5238617, 1570410, 5650371}.



Contract-based modeling~\cite{benveniste2018contracts,sangiovanni2012taming,damm2011using,Nuzzo14,Nuzzo14_1,Nuzzo15b,nuzzo2018chase} can be suited to formalize and analyze properties of reactive systems. A \textit{contract} specifies the behavior of a component by distinguishing its responsibilities  (\textit{guarantees}) from those of its environment (\textit{assumptions}). It is possible to use contracts to model the mission specification and automatically realize its implementation using reactive synthesis~\cite{mallozzi2020crome}.


However, reactive synthesis is impractical for large specifications due to its high computational complexity (double exponential in the length of the formula). Breaking the specification into more manageable chunks that can be realized independently can reduce the computational complexity. Alternatively, instead of generating the implementation, we can search for efficient implementations that can be composed and together realize the specification. It is then essential to decouple the specification of a generic robotic mission from its possible implementations. Different implementations can, for example, refer to different robotic systems on which the mission will be deployed.

\pierg{formalizing requirement to specification first step. Not final solution form the beginning. Adaptation. Mapping to library second step. }

\pierg{TODO: Furthermore: connecting abstraction layers, PBD..?}

System specifications formalized with contracts can be incrementally refined using a \textit{library} of pre-defined components~\cite{iannopollo2014library,mallozzi2020cogomo}. 
In the context of robotic missions, a component is a pre-implemented mission specification. A library containing such components can be used to find suitable refinements of the mission specification. Using a library, we can also \textit{adapt} the specification to a variety of possible implementations. \ins{For example, a general `search and rescue' mission can be automatically adapted to be deployed in different environments (e.g., different map configurations).}
Ideally, the refinement process should use elements in the library of pre-implemented specifications~\cite{sangiovanni2012taming}. Since every component of the library can be pre-implemented, we would not need to generate implementations for every specification, but we can \textit{reuse} existing ones. Furthermore, various libraries can model different robotic systems or system aspects. Each library can add additional constraints that the specification must meet to be implemented.
However, the designer might not be aware \emph{a priori} of the library's use, or the library may not be rich enough to ``cover'' the specification. The question is then how to add the minimum number of components to the library to refine the specification completely. 

In this paper, we present \NAME: a structured methodology to \ins{model}, search and repair formal specifications represented as assume-guarantee contracts~\cite{benveniste2008}. The search process consists in keeping the specification fixed while searching for the \textit{missing part} in other libraries. The repairing process consists of automatically \textit{patching} the current specification such that the available library can refine it. Whenever a specification cannot be refined from the library, we propose an algorithm to produce the best `candidate selection' of elements such that the `missing part' to search or repair is minimal \ins{(with respect to the number of behaviors)}, maximizing the available library's use.
%

The contributions of this paper are the following:
\begin{itemize}
\item \ins{a framework to model mission specification and to prove specification refinements across various abstraction layers.}
 \item an algorithm to find the best candidate selection of library elements based on syntactic and semantic similarities with the specification to be refined.
 \item a methodology to search or repair specifications that cannot be refined. The search is based on the application of the contract operations of \textit{quotient} \cite{agquotient} and \textit{composition} \cite{benveniste2018} while the repair is based on the operations of \textit{separation} and \textit{merging} \cite{passerone2019coherent}.
\end{itemize}

\ins{We have implemented \NAME in a tool that supports the designer in the mission specification modeling, refinement, and repairing process\footnote{Tool available: \textcolor{blue}{hidden for blind review}}.}

\paragraph{Related Works.}

Repairing of system specifications is a widely studied problem in the literature. Recent approaches have focused on repairing LTL specifications that are unrealizable for reactive synthesis~\cite{alur2013counter, cavezza2017interpolation, chatterjee2008environment, maoz2019symbolic, li2011mining}. These approaches focus on the discovery of \textit{new assumptions} to make the specification realizable. They use a restricted fragment of LTL specifications (e.g., GR(1)) and mostly use model checking techniques. More recent approaches by Gaaloul et al.~\cite{gaaloul2020mining} rely on testing rather than model checking to generate the data used to learn assumptions using machine learning techniques, and apply them to complex signal-based modeling notations rather than to LTL specifications. 
Brizio et al.~\cite{brizzio2021automated} use a search-based approach to repair LTL specifications. Their approach is based on syntactic and semantic similarity, where their heuristic is based on \textit{model counting}, i.e., the number of models that satisfy the formula. The new realizable specification is then produced by successive application of genetic operations. 
For repairing Signal Temporal Logic (STL) formulas, Gosh et al.~\cite{ghosh2016diagnosis} propose algorithms to detect possible reasons for infeasibility and suggest repairs to make it realizable. Approaches such as~\cite{ergurtuna2021automated, chatzieleftheriou2012abstract}, instead of repairing the specification, focus on repairing the \textit{system} in a way that it can satisfy the specification.
In the robotics domain, Boteanu et al.~\cite{pacheck2019automatic} focus on adding assumptions to the robot specification while having the human prompted to confirm or reject them. Pacheck et al.~\cite{pacheck2019automatic, pacheck2020finding} automatically encode into LTL formulas robot capabilities based on sensor data. If a task cannot be performed (i.e., the specification is unrealizable), they suggest skills that would enable the robot to complete the task.

\pierg{TODO: add other related for library / missing component}

Our framework uses LTL specifications, but instead of mining for new assumptions or changing the system (i.e., the pre-implemented library goals), it repairs existing specifications (assumptions and guarantees) based on what can be refined from the library of goals. Our repairs are always the smallest (in terms of the behaviors removed from the original specification) and completely automated since they are based on algebraic operations. We use ideas similar to~\cite{brizzio2021automated} to compute the candidate composition, which is based on semantic and syntactic similarities to the goal to be refined.

\section{Background}\label{sec:background}
We provide some background on assume-guarantee contracts and linear temporal logic.

\subsection{Assume-Guarantee Contracts}
\label{sec:backgroundcontracts}

Contract-based design~\cite{benveniste2018contracts,Nuzzo15b} has emerged as a design paradigm capable of providing formal support for building complex systems in a modular way by enabling compositional reasoning, step-wise refinement of specifications, and reuse of pre-designed components. 

A \emph{contract} $\mathcal{C}$ is a triple $(V, A, G)$ where $V$ is a set of system \textit{variables} (including, e.g., input and output variables or ports), and $A$ and $G$---the assumptions and guarantees---are sets of behaviors over $V$. For simplicity, whenever possible, we drop $V$ from the definition and refer to contracts as pairs of assumptions and guarantees, i.e., $\mathcal{C}=(A, G)$. $A$ expresses the behaviors expected from the environment, while $G$ expresses the behaviors that an implementation promises under the environment assumptions. In this paper, we express assumptions and guarantees as sets of behaviors satisfying a logical formula; we then use the formula itself to denote them. An environment $E$ satisfies a contract $\mathcal{C}$ whenever $E$ and $\mathcal{C}$ are defined over the same set of variables, and all the behaviors of $E$ are included in the assumptions of $\mathcal{C}$, i.e., when $|E| \subseteq A$, where $|E|$ is the set of behaviors of $E$. An implementation $M$ satisfies a contract $\mathcal{C}$ whenever $M$ and $\mathcal{C}$ are defined over the same set of variables, and all the behaviors of $M$ are included in the guarantees of $\mathcal{C}$ when considered in the context of the assumptions $A$, i.e., when $|M| \cap A \subseteq G$. 

A contract $\mathcal{C}=(A, G)$ can be placed in saturated form by re-defining its guarantees as $G_{sat} = G \cup \overline{A}$, where $\overline A$ denotes the complement of $A$. A contract and its saturated form are semantically equivalent, i.e., they have the same set of environments and implementations. 
Therefore, in the rest of the paper, we assume that all the contracts are expressed in saturated form. In particular, the relations and operations we will discuss are only defined for contracts in saturated form.
A contract $\mathcal{C}$ is \emph{compatible} if there exists an environment for it, i.e., if and only if $A \neq \emptyset$. Similarly, a saturated contract $\mathcal{C}$ is \emph{consistent} if and only if there exists an implementation satisfying it, i.e., if and only if $G \neq \emptyset$. We say that a contract is \emph{well-formed} if and only if it is compatible and consistent. We detail below the contract operations and relations used in this paper. 

\paragraph{Contract Refinement.} 
Refinement establishes a pre-order between contracts, which formalizes the notion of replacement. Let
$\mathcal{C} = (A, G)$ and $\mathcal{C}' = (A', G')$ be two contracts, we say that $\mathcal{C}$ refines $\mathcal{C}'$, denoted by  $\mathcal{C} \preceq \mathcal{C}'$, if and only if all the assumptions of $\mathcal{C}'$ are contained in the assumptions of $\mathcal{C}$ and all the guarantees of $\mathcal{C}$ are included in the guarantees of $\mathcal{C}'$, that is, if and only if  
$A \supseteq A' \text{ and } G \subseteq G'$.
Refinement entails relaxing the assumptions and strengthening the guarantees. When $\mathcal{C} \preceq \mathcal{C}'$, we also say that $\mathcal{C}'$ is an \textit{abstraction} of $\mathcal{C}$ and can be replaced by $\mathcal{C}$ in the design.

\paragraph{Contract Composition.}
The operation of composition ($\parallel$) is used to generate the specification of a system made of components that adhere to the contracts being composed. 
Let $\mathcal{C}_1 = (A_1, G_1)$ and $\mathcal{C}_2 = (A_2, G_2)$ be two contracts. The composition $\mathcal{C}=(A,G)=\mathcal{C}_1  \parallel  \mathcal{C}_2$ can be computed as follows:
\begin{align}
 A & = (A_1 \cap A_2) \cup \overline{(G_1 \cap G_2)}, \label{eq:sat_composition_A}\\
 G & = G_1 \cap G_2. \label{eq:sat_composition_G}
\end{align}
Intuitively, an implementation satisfying $\mathcal{C}$ must satisfy the guarantees of both $\mathcal{C}_1$ and $\mathcal{C}_2$, hence the operation of intersection in~\eqref{eq:sat_composition_G}.
An environment for $\mathcal{C}$ should also satisfy all the assumptions, motivating the conjunction of $A_1$ and $A_2$ in~\eqref{eq:sat_composition_A}. However, part of the  assumptions in $\mathcal{C}_1$ may be already supported by $\mathcal{C}_2$ and \emph{vice versa}. This allows relaxing $A_1 \cap A_2$ with the complement of the guarantees of $\mathcal{C}$~\cite{benveniste2018contracts}.

\paragraph{Quotient (or Residual).}
Given two contracts $\mathcal{C}_1$ and $\mathcal{C}'$, the quotient $\mathcal{C}_2 =(A_2, G_2) = \mathcal{C}' / \mathcal{C}_1$, is defined as the largest specification that we can compose with $\mathcal{C}_1$ so that the result refines $\mathcal{C}'$. In other words, the quotient is used to find the specifications of missing components.
We can compute the quotient \cite{agquotient} as follows: 
\begin{align*}
    A_2 & = A' \cap G_1\quad \text{and} \quad
    G_2  = G' \cap A_1 \cup \overline{(A' \cap G_1)}.
\end{align*}
%
%
\paragraph{Contract Merging.}
Contracts that handle specifications of various viewpoints of the same design element can be combined using the \textit{merging} operation \cite{passerone2019coherent}. Given  $\mathcal{C}_1 = (A_1, G_1)$ and $\mathcal{C}_2 = (A_2, G_2)$ their \textit{merger} contract, denoted $\mathcal{C} = \mathcal{C}_1  \bigcdot  \mathcal{C}_2$, is the contract which promises the guarantees of both specifications when the assumptions of both specifications are respected, that is,
\begin{equation*}
    \mathcal{C} = (A_1 \cap A_2, G_1 \cap G_2 \cup \overline{A_1 \cap A_2}). 
\end{equation*}
\paragraph{Separation.}
Given two contracts $\mathcal{C}_1$ and $\mathcal{C}'$ the operation of \textit{separation} \cite{passerone2019coherent} computes the contract $\mathcal{C}_2 =(A_2, G_2) = \mathcal{C}' \div \mathcal{C}_1$ as
\begin{align*}
    A_2 & = A' \cap G_1 \cup \overline{(G' \cap A_1)} \quad\text{and}\quad
    G_2  = G' \cap A_1
\end{align*}
The contract $\mathcal{C}_2$ is defined as the smallest
(with respect to the refinement order)
contract satisfying
$
  \mathcal{C}' \preceq \mathcal{C}_1 \bigcdot \mathcal{C}_2
$.
\subsection{Linear Temporal Logic}

Given a set of atomic propositions $AP$ (i.e., Boolean statements over system variables) and the state $s$ of a system (i.e., a specific valuation of the system variables), we say that \textit{$s$ satisfies $p$}, written $s \models p$, with $p \in AP$, if $p$ is $\textit{true}$ at state $s$.
We can construct LTL formulas over $AP$ according to the following recursive grammar:
\begin{align*}
    \varphi := p ~|~ \neg\varphi ~|~ \varphi_1 \lor \varphi_2 ~|~ \textbf{X} ~\varphi ~|~ \varphi_1 ~\textbf{U}~ \varphi_2
\end{align*}
where $\varphi$, $\varphi_1$, and $\varphi_2$ are LTL formulas.
From the negation ($\neg$) and disjunction $(\lor)$ of the formula we can define the conjunction ($\land$), implication ($\rightarrow$), and equivalence ($\leftrightarrow$). Boolean constants \textit{true} and \textit{false} are defined as $\textit{true}= \varphi \lor \neg \varphi$ and $\textit{false} = \neg \textit{true}$. 
The temporal operator $\textbf{X}$ stands for \textit{next} and $\textbf{U}$ for \textit{until}. 
Other temporal operators such as \textit{globally} ($\textbf{G}$) and \textit{eventually} ($\textbf{F}$) can be derived as follows: $\textbf{F}~ \varphi = \textit{true} ~\textbf{U}~ \varphi$ and $\textbf{G} ~ \varphi = \neg(\textbf{F}(\neg \varphi))$.
We refer to the literature~\cite{baier2008principles} for the formal semantics of LTL. For the rest of the paper, we indicate with $\overline{\varphi}$ the negation of $\varphi$, i.e. $\neg \varphi$.

\paragraph{Reactive Synthesis}
An LTL formula can be `\textit{realized}' into a controller via reactive synthesis~\cite{F16}.
Reactive synthesis generates a controller $\mathcal{M}$ (a finite state machine) from a specification $\varphi$ (an LTL formula) having its atomic propositions divided into inputs and outputs. If a controller can be produced, it is guaranteed to satisfy the specification under all possible inputs. If such machine exists, we say that $\mathcal{M}$ \textit{realizes} $\varphi$.

\section{Problem Definition}
\label{sec:overview}
Robotic missions state what the robot should achieve in the world. We model each robot objective with a \textit{contract}.





\begin{definition}[Mission Specification]
  A mission specification is a contract $\mathcal{C} = (\varphi_A, \varphi_G)$ of LTL specifications, where $\varphi_A$ defines the behaviors assumed of the environment and $\varphi_G$ the behaviors the robot is allowed when the environment meets the assumptions. 
\end{definition}

A behavior is an infinite sequence of states, where each state is an assignment of values to all system variables within their domain.
\ins{A finite state machine is a tuple $\mathcal{M}=(S, \mathcal{I}, \mathcal{O}, s_0, \delta)$ where $S$ is the set of states, $s_0 \in S$ is the initial state, and $\delta:S \times 2^\mathcal{I} \rightarrow S \times 2^\mathcal{O}$ is the transition function.
}A finite state machine $\mathcal{M}$
realizes a contract $\mathcal{C} = (\varphi_A, \varphi_G)$, denoted $\mathcal{M} \models \mathcal{C}$, if it realizes the formula $\varphi = \varphi_A \rightarrow \varphi_G$.




\begin{definition}[Library of Components]
  A library of components is a pair $\Delta=(K, M)$, where $K=\{\mathcal{L}'_1, \mathcal{L}'_2, \dots , \mathcal{L}'_n\}$ is a set of $n$ contracts and $M=\{\mathcal{M}'_1, \mathcal{M}'_2, \dots , \mathcal{M}'_n\}$ is a set of $n$ finite state machines such that $\mathcal{M}'_i \models \mathcal{L}'_i$ for all $i$.
\end{definition}


The library of components \textit{bridges the gap} between a general specification and a specific set of implementations that can be executed in a certain environment. 
The robot is a finite state machine that satisfies the mission specification using the library of components.

\begin{definition}[Mission Satisfaction Problem]\label{prob:prob}
Given a mission specification $\mathcal{C}$ and a library of components $\Delta=(K, M)$, produce an implementation $\mathcal{M} = \mathcal{M}_1 ~\parallel~ \mathcal{M}_2 ~\parallel~ \dots ~\parallel~ \mathcal{M}_p$ where $\{\mathcal{M}_1, \mathcal{M}_2, \dots, \mathcal{M}_p\} \in M$ such that $\mathcal{M} \models \mathcal{C}$.
\end{definition}

However, we cannot always find components in the library that can satisfy the mission specification, e.g., the library does not `\textit{cover}' all the constraints of the mission specification or does not support parts of the specification. 
When $\mathcal{M} \not\models \mathcal{C}$, we propose two strategies: \textit{1)} loosening the specification $\mathcal{C}$ by relaxing its constraints or \textit{2)} extending $M$ with new components that can accommodate the constraints imposed by $\mathcal{C}$. 
Our framework, named \NAME, automatically performs both strategies while satisfying optimality criteria by leveraging the contract algebra operations.


\section{Running Example}
\label{sec:example}

\begin{figure}[h]
  \centering
  \includegraphics[width=0.5\columnwidth]{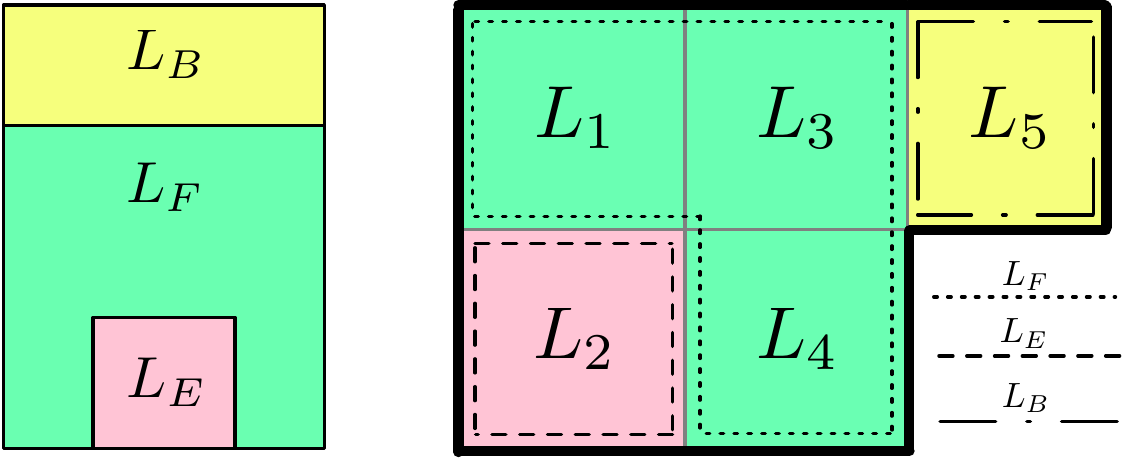}
  \caption{Running Example Location Maps. The map on the left of the picture is `refined' by the map on the right.}
  \label{fig:example}
\end{figure}

Let us consider an environment modeling a general store formed by a front, a back, and an entrance. The left part of Figure~\ref{fig:example} shows such a map with three locations: $L_F$ (\textit{front}), $L_B$ (\textit{back}), and $L_E$ (\textit{entrance}). We have that location $L_E$ is inside $L_F$. We assume the robot is equipped with a sensor to detect people and actuators to greet them. The mission consists of moving between the back and the front of the store and greeting customers when they are detected.

Let us assume we want to deploy the mission on a specific store with the map shown on the right side of Figure~\ref{fig:example}. Here we have five locations $L_1, \ldots, L_5$, which indicate where the robot can be. We have that locations $L_1, L_3$ and $L_4$ represent the front of the store, location $L_2$ the entrance, and location $L_5$ the back. Let us assume that we have a library of components $\Delta$ containing implementations of mission objectives that the robot can perform in this more detailed map.

Our goal is to \textit{refine} a general mission specification with a set of implementations specific to a particular environment. In the rest of the paper, we will see examples of how \NAME tackles this problem, even in cases when the refinement is not possible at first.

\section{Modeling and Well-formedness}
\label{sec:modeling}
We now introduce the building blocks of the modeling infrastructure used in \NAME, starting with the concept of \textit{types}.
Types are used to assign \emph{semantics} to every location, sensor, and action in the mission. Then relationships among types are used to automatically generate constraints to model the world in which the robot operates. We call \emph{world context} the ground constraints that model the world.


\begin{definition}[Type]
    A type is a semantic concept related to the mission (e.g., a location, an action, or a sensor) that is used to generate the world context. We indicate with $\Theta$ the set of all types in scope.
\end{definition}

Every type comes with one atomic proposition, i.e., a variable that can have two values: \textit{true} and \textit{false}. We will indicate types with capital letters and atomic propositions with their corresponding lower-case letter.

Types can be related to other types in four ways: mutual exclusion, adjacency, extension (or subtyping), and covering.

\begin{definition}[Mutual exclusion]
    A type $A \in \Theta$ is mutually exclusive from a type $B \in \Theta$ if instances of $A$ and $B$ can never be \textit{true} simultaneously.
\end{definition}

This relationship will be used to state that the robot cannot be in two locations simultaneously.

\begin{definition}[Adjacency]
    A type $A \in \Theta$ is adjacent to a type $B \in \Theta$ if instances of $B$ can become \textit{true} one step after instances of $A$ are \textit{true}.
\end{definition}

This relationship will help us specify that the reachable locations from a given location in one timestep are those allowed by the current map.

\begin{definition}[Extension]
    Let $A \in \Theta$ and $B \in \Theta$ be two types, where $A \neq B$. We say that $A$ is a \textit{subtype} of $B$ or that $A$ \textit{extends} $B$ if the concept $A$ is included in the concept $B$.
    We denote extension among types as $A \preceq B$.
\end{definition}

Subtyping is used to relate abstract to concrete types. For example, in Figure \ref{fig:example}, we will define a type $L_F$ denoting ``the robot is in the front of the store'' and another denoting ``the robot is in location $L_4$.'' Because $L_4$ is part of the front, we will say that $L_4$ is a subtype of $L_F$.

\begin{definition}[Covering]
    Let $A \in \Theta$ be a type and $A_i \in \Theta$ ($i \le n$) be subtypes of $A$. We say that the set $\{A_1', \ldots, A_n'\}$ covers the type $A$ if, when an atomic proposition of $A$ is true, the atomic propositions of at least one of the $A_i'$ are true.
\end{definition}

The concept of covering allow us to say that an abstract type is represented exactly by a disjunction of concrete types. For example, in Figure \ref{fig:example}, we say that the type ``the robot is in the front'' is covered by the set $\{L_1, L_3, L_4\}$ since to be in the front requires the robot to be in one of those specific locations.

\begin{definition}[Similar types]\label{def:similartypes}
    A type $A \in \Theta$ is \textit{similar} to a type $B \in \Theta$ iff $A \preceq B$ or $A = B$.
\end{definition}


Our modeling framework uses types to generate world context constraints semantically.
For each type of relationship described above, our framework produces an LTL formula that can be added to the world context.
We refer the reader to Appendix~\ref{sec:appendix} to learn how \NAME generates the context constraints and verifies \textit{consistency}, \textit{refinement} and \textit{realizability} of specifications.

\begin{example}
We consider the mission specification of our running example to be the following contract $\mathcal{C}$:
\begin{align}\label{spec:c}
\mathcal{C}
      \begin{cases}
        A & \mathsf{InfOften}(p)\\
        G & \mathsf{OrderedPatrolling}(l_b, l_f) ~ \land 
          ~ \mathsf{InstantaneousReaction}(s, g)
      \end{cases}
\end{align}

$\mathsf{InfOften}$ represents the LTL construct to express \textit{globally eventually} ($\mathsf{InfOften}$);   
$\mathsf{OrderedPatrolling}$ is a robotic pattern \ins{(i.e. template for a robotic specification)} that express the continuous visit of a set of locations imposing an order during the visit. 
$\mathsf{InstantaneousReaction}$ is another robotic pattern which in the same time step performs an action (i.e., sets its atomic proposition to \textit{true}, i.e. $g$, based on the truth value of a different atomic proposition, i.e. $s$). 
For more details and the complete list of robotic patterns, see Menghi et al.~\cite{menghipatterns}.
\end{example}

\begin{figure*}[t]
    \centering
    \includegraphics[width=0.8\linewidth]{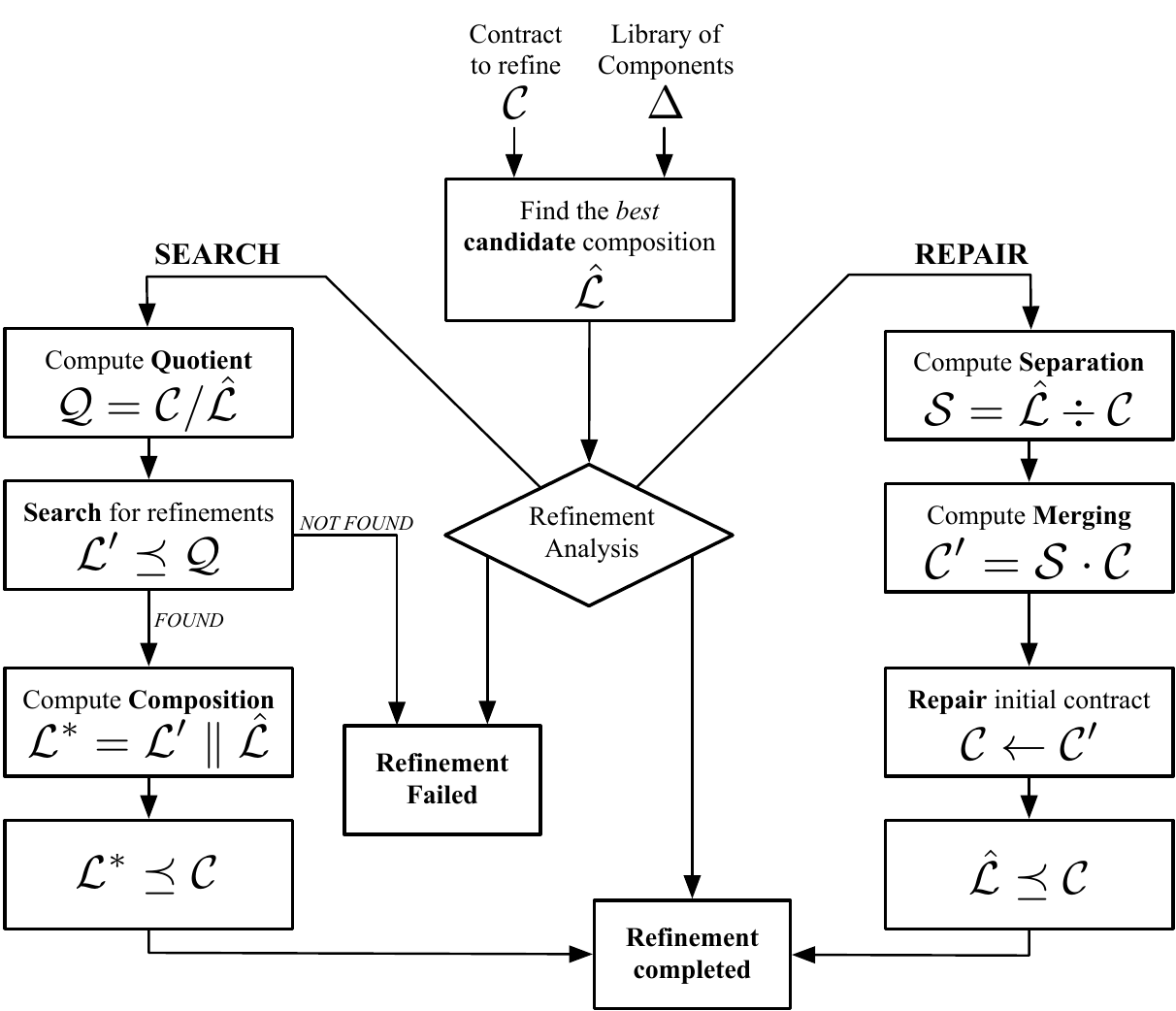}
    \caption{Flow diagram showing all the processes involved in creating a refinement from a library.}
    \label{fig:flowdiagram}
   \end{figure*}

\section{Contract Search and Repair}
\label{sec:refandrep}

Figure~\ref{fig:flowdiagram} shows all the processes involved in the refinement of a contract $\mathcal{C}$ using a library of components $\Delta$. First, \NAME searches for the best \textit{candidate composition} of contracts $\hat{\mathcal{L}}$ from $\Delta$, i.e., the selection of contracts that once composed maximize some heuristic function related to the refinement of $\mathcal{C}$. Then, according to the result of a \textit{refinement analysis} procedure, \NAME can either declare the refinement \textit{complete}, start a \textit{search} procedure, start a \textit{repair} procedure, or declare the refinement \textit{failed}. In the search process we look for new contracts in order to be able to refine $\mathcal{C}$, whereas in the repair process we automatically modify $\mathcal{C}$ such that $\Delta$ can refine it.
In the following sections, we discuss \textit{1)} how \NAME produces an optimal candidate composition, \textit{2)} the refinement analysis procedure, \textit{3)} the search process using the contract operations of quotient and composition, \textit{4)} the repair process using the contract operations of separation and merging.

\subsection{Finding the Best Candidate Composition}
\label{sec:findbestcandidate}
The best candidate composition $\hat{\mathcal{L}}$ is a composition of a selection of contracts in the library that aims to be \textit{`the closest refinement'} of $\mathcal{C}$ that can be generated from $\Delta$. We define the closest refinement by formulating a heuristic function $h$. Thus, given $\mathcal{C}$ and $\Delta$, the best candidate composition is the composition of contracts in $\Delta$ that maximizes $h$. The heuristic function $h$ is based on \textit{1)} type similarity, and \textit{2)} behavior coverage.

Let $\Gamma$ be the class of all contracts and $\Theta$ a set of all types. We define the following functions:

\begin{definition}[Similarity Score]
 $SIM: \Gamma \times 2^\Theta \rightarrow \mathbb{N}$ is a function that takes as input a contract $\mathcal{C} \in \Gamma$ and a set of types $\Theta_i \in 2^\Theta$ and returns the number of \textit{similar types} (\ref{def:similartypes}) between the types of the specification in $\mathcal{C}$ and $\Theta_i$. Let $\Theta_i$ be the set of types of the contracts in $\Gamma_i \in 2^\Gamma$ and $N$ be the number of types of a contract $\mathcal{C} \in \Gamma$. The \textit{similarity score} of a set of contracts $\Gamma_i$ with respect to $\mathcal{C}$ is
 \begin{equation}
     \frac{SIM(\mathcal{C}, \Theta_i)}{N} \times 100.
 \end{equation}
That is the percentage of similar types `covered' by $\Gamma_i$.
\end{definition}

\begin{definition}[Refinement Score]
The \textit{refinement score} of contract $\mathcal{C}$ with respect to a set of contracts $\Gamma_i \in 2^\Gamma$ is a function $REF: \Gamma \times 2^\Gamma \rightarrow \mathbb{R}$ that returns percentage of contracts in $\Gamma_i$ that can be refined by $\mathcal{C}$.
\end{definition}

We can assign a refinement score to every contract in the library by making a \textit{pair-wise comparison} among all selections of contracts in the library. This process can be done separately from searching for the best candidate composition.

Our framework starts the search for the best candidate composition by looking at the possible combinations of contracts that are composable and computing for each of them their similarity score with respect to $\mathcal{C}$.
Let $\Omega$ be the set compositions with the highest similarity score.
If $\Omega$ has more than one element, i.e., $|\Omega| > 1$, then it keeps in $\Omega$ only the compositions generated by the \textit{least number of contracts}. Then, if $|\Omega| > 1$, we compute the refinement score of the compositions in $\Omega$ by making a pair-wise comparison among them if their score has not been computed offline already. The best candidate composition is the element in $\Omega$ having the highest refinement score. We choose one randomly if more than one element has the highest score.

\begin{example}

Let us consider a simplified version of $\mathcal{C}$ in (\ref{spec:c}) that only contains $\mathsf{OrderedPatrolling}(l_f, l_b)$ as a guarantee which corresponds to the contract $\mathcal{C}_1$:



\begin{align}\label{spec:c1}
    \mathcal{C}_1
          \begin{cases}
            A & \textit{true}\\
            G & \mathsf{G} \mathsf{F} (\mathit{l_f} \land \mathsf{F} \mathit{l_b}) \land (\overline{\mathit{l_b}} \mathbin{\mathsf{U}} \mathit{l_f}) ~ 
            \land~ \mathsf{G} (\mathit{l_b} \rightarrow \mathsf{X} (\overline{\mathit{l_b}} \mathbin{\mathsf{U}} \mathit{l_f})) \land \mathsf{G} (\mathit{l_f} \rightarrow \mathsf{X} (\overline{\mathit{l_f}} \mathbin{\mathsf{U}} \mathit{l_b}))
          \end{cases}
\end{align}

The contract in (\ref{spec:c1}) imposes a continuous visit of locations $l_f, l_b$, (i.e., their atomic propositions must be infinitely often \textit{true}). Furthermore, it imposes that the locations must be visited in order starting from $l_f$.

Let us assume that our library of contracts $\Delta=(K, M)$ has $K=\{\mathcal{L}_1, \mathcal{L}_2, \mathcal{L}_3, \mathcal{L}_4\}$:
\vspace{-2mm}

\begin{align}
 &\mathcal{L}_{1} 
 \begin{cases}
 G & \mathsf{Patrolling}(l_5)\\
  & \mathsf{G} \mathsf{F} l_{5}
 \end{cases} 
 &&\mathcal{L}_{2} 
 \begin{cases}
 G & \mathsf{Patrolling}(l_3)\\
  & \mathsf{G} \mathsf{F} l_{3}
 \end{cases} \notag\\
 & \mathcal{L}_{3} 
 \begin{cases}
 G & \mathsf{Visit}(l_3, l_1)\\
  & \mathsf{F} l_{3} \land \mathsf{F} l_{1}
 \end{cases} 
 && \mathcal{L}_{4} 
 \begin{cases}
 G & \mathsf{Visit}(l_5)\\
  & \mathsf{F} l_{5}
 \end{cases} \notag
 \end{align}
 

When not differently stated, we consider the assumptions to be \textit{true}.
In the case of $\Delta$ represented above, we have indicated both the robotic pattern and its LTL representation for the guarantees of each contract.

Let us compute the composition of contracts in $K$ that has the best similarity and refinement score.
Seven candidates have the best similarity score (i.e., $100\%$) between $\mathcal{C}$ and all the types in $\Delta$. Among these, three results from the composition of two contracts, and the rest are composed of more than three contracts. After filtering out the candidates formed by the composition of more than two contracts, \NAME chooses the candidate with the highest refinement score, which is, in this case, is $\hat{\mathcal{L}} = \mathcal{L}_1 \parallel \mathcal{L}_2$ and the resulting contract has the following guarantees:
\begin{equation}
 \mathsf{G} \mathsf{F} l_{5} \land \mathsf{G} \mathsf{F} l_{3} \label{eq:guaranteescandidate}
\end{equation}
We can see how we have found the composition of contracts in $\Delta$ that generates behaviors that are the most `similar' to those of the contract that we want to refine (\ref{spec:c1}).
Choosing the most refined combination of contracts allows us to maximize our contracts' usage in the library. Specifically, we can maximize the number of behaviors of $\mathcal{C}$ that can be \textit{covered} by $\Delta$.

\end{example}

\subsection{Refinement Analysis}
The refinement analysis evaluates the best candidate composition $\hat{\mathcal{L}}$ and determines the appropriate strategy to complete the refinement of $\mathcal{C}$ from $\Delta$. The outcome of the analysis can be one of the following:

\begin{itemize}
 \item \textit{Refinement Failed}: $\mathcal{C}$ can not be refined by $\Delta$.
 \item \textit{Refinement Completed}: $\hat{\mathcal{L}}$ is already a refinement of $\mathcal{C}$, i.e., $\hat{\mathcal{L}} \preceq \Delta$.
 \item \textit{Start Search Procedure}: $\hat{\mathcal{L}} \not\preceq \Delta$, start the search procedure for a new specification using contract quotient and composition.
 \item \textit{Start Repair Procedure}: $\hat{\mathcal{L}} \not\preceq \Delta$, start a repair procedure to modify $\hat{\mathcal{C}}$ such that $\Delta$ can refine it. This process uses contract merging and separation.
\end{itemize}

Algorithm~\ref{alg:analysis} shows the main steps of the refinement analysis procedure. In addition to $\mathcal{C}$ and $\Delta$, the algorithm can take as input additional libraries $D = \{\Delta' \dots \Delta'' \dots \}$. Moreover, users can express their intention of performing a \textit{repair} or a \textit{search} procedure.
If the designers do not express procedure preferences, \NAME chooses a procedure based on the similarity score. If the library \textit{covers} all the types of $\mathcal{C}$ \ins{or its similarity score is at least $80\%$,} the algorithm performs a repair of the specification. Otherwise, other types can likely be found in the additional libraries if provided by the designer. Hence, if different libraries are available and the similarity score is less than $80\%$, \NAME performs the search procedure. The following two sections describe the search and repair procedures and illustrate how they are applied to our running example.

\begin{algorithm}[h]
  \DontPrintSemicolon
  {\footnotesize 
    
  \KwInput{$\mathcal{C}$: contract to refine, $\Delta$: library of components, $\hat{\mathcal{L}}$: best candidate composition, $D = \{\Delta' \dots \Delta'' \dots \}$: additional libraries of components (\textit{optional}), \texttt{repair}, \texttt{search}: Boolean arguments indicating the designer preference (\textit{optional})}
  \KwOutput{\texttt{refinement\_complete}: Boolean indicating that the refinement procedure has been completed, \texttt{refinement}: contract refining $\mathcal{C}$}

  \If{$\hat{\mathcal{L}} \preceq \Delta$}
  {
      \tcc{$\hat{\mathcal{L}}$ is already a refinement of $\Delta$}
      \textbf{return} \textit{true}, $\hat{\mathcal{L}}$
  }
  \If{$ \textit{similarity score} == 0\%$}
  {
      \tcc{Refinement failed}
      \textbf{return} \textit{false}, \textit{None}
  }
  \If{\texttt{repair}}
  {
      \tcc{Designer is `forcing' a repair}
      \textbf{return} $\textbf{repair\_procedure}(\hat{\mathcal{L}}, \mathcal{C})$
  }
  \If{$\texttt{search} \land L \neq \emptyset$}
  {
      \tcc{Designer is `forcing' a search}
      \textbf{return} $\textbf{search\_procedure}(\hat{\mathcal{L}}, \mathcal{C}, D)$
  }
  \tcc{Choose Search or Repair based on the similarity score}
  \If{$ \textit{similarity score} \ge 80\%$}
  {
    \textbf{return} $\textbf{repair\_procedure}(\hat{\mathcal{L}}, \mathcal{C})$
    }
  \If{$L \neq \emptyset$}
    {
        \tcc{Designer is `forcing' a search}
        \textbf{return} $\textbf{search\_procedure}(\hat{\mathcal{L}}, \mathcal{C}, D)$
    }
    \textbf{return} \textit{false}, \textit{None}

  }
  \caption{Refinement Analysis}
  \label{alg:analysis}
  \end{algorithm}

\subsection{Specification Search via Quotient and Composition}
\label{sec:quotient}
We have seen in Section~\ref{sec:findbestcandidate} that the best candidate composition $\hat{\mathcal{L}}$ is the most refined composition of contracts. It means that we have \textit{delegated} as much functionality as possible to the library of contracts $\Delta$.
We need to find the specification that $\Delta$ cannot meet but that we still need to satisfy to refine $\mathcal{C}$. 

To find this \textit{missing part} given $\hat{\mathcal{L}}$ and $\mathcal{C}$, we would like to have a specification that is as general as possible. The contract operation of quotient suits our needs perfectly, as it produces the most abstract specification that, composed with $\hat{\mathcal{L}}$, can refine $\mathcal{C}$. Then \textit{any refinement} of the quotient can be substituted in the composition, and we still obtain a refinement of $\mathcal{C}$.

As shown in Figure~\ref{fig:flowdiagram}, we first compute the \textit{quotient} between $\mathcal{C}$ and $\hat{\mathcal{L}}$, i.e., $\mathcal{Q} = \mathcal{C} / \hat{\mathcal{L}}$. 
Then we refine the quotient by searching for new specifications $\mathcal{L}'$ such that $\mathcal{L}' \preceq \mathcal{Q}$. The refinement $\mathcal{L}'$ can be searched in a a library of contracts $\Delta' \in D$. The \textit{search\_procedure} will search in all the libraries in $D$, and once a refinement of the quotient is found, we compose it with $\hat{\mathcal{L}}$, i.e., $\mathcal{L}^* = \mathcal{L}' \parallel \hat{\mathcal{L}_i}$. The resulting contract $\mathcal{L}^*$ is guaranteed to refine $\mathcal{C}$.

If there is no refinement of the quotient in any of the libraries in $D$, then the refinement process fails. At this point, the designer could choose to \textit{delegate} to some third-party the implementation of the quotient by giving them $\mathcal{Q}$.



\begin{example}

Let us continue with the example in the previous section, where we found that the best candidate composition of $\mathcal{C}_1$ in (\ref{spec:c1}) using the library $\Delta$ is $\hat{\mathcal{L}} = \mathcal{L}_1 \parallel \mathcal{L}_2$. 
   Even though the similarity score is maximum, let us see what happens if the designer imposes the search for new contracts providing a new library, i.e., $D=\{\Delta'\}$.
   \pierg{rev: what is contained in Delta'?}
   
   We can compute the quotient $\mathcal{Q} = \mathcal{C}_1 / \hat{\mathcal{L}}$, which has the following contract:

\begin{equation}
  \mathcal{Q} =
    \begin{cases}
      A & \mathsf{G} \mathsf{F} l_{5} \land \mathsf{G} \mathsf{F} l_3\\
      G & (\mathsf{G} \mathsf{F} (\mathit{l_f} \land \mathsf{F} \mathit{l_b}) \land (\overline{\mathit{l_b}} \mathbin{\mathsf{U}} \mathit{l_f}) ~~~\land \\
      & \land ~~~ \mathsf{G} (\mathit{l_b} \rightarrow \mathsf{X} (\overline{\mathit{l_b}} \mathbin{\mathsf{U}} \mathit{l_f})) \land \mathsf{G} (\mathit{l_f} \rightarrow \mathsf{X} (\overline{\mathit{l_f}} \mathbin{\mathsf{U}} \mathit{l_b}))) 
      ~~ \lor ~~~ \overline{\mathsf{G} \mathsf{F} l_{5} \land \mathsf{G} \mathsf{F} l_3}
    \end{cases}
\end{equation}

The quotient is the result of an algebraic expression and is computed automatically; without looking at the specifications, the designer knows the missing behavior from library $\Delta$ such that $\mathcal{C}$ can be refined.
In fact, any refinement of $\mathcal{Q}$ can serve to `complete' the candidate composition $\hat{\mathcal{L}}$ so that it refines $\mathcal{C}$. \NAME searches for refinements of $\mathcal{Q}$ from $\Delta'$ to produce a new candidate composition. Let $\mathcal{L}'$ in (\ref{spec:lprime}) be the candidate composition that completely refines $\mathcal{Q}$. $\mathcal{L}'$ indicates a strict order among locations to be visited, similar to the \textsf{StrictOrderedPatrolling} robotic patterns, but without prescribing that they be continuously visited.


\begin{align}\label{spec:lprime}
    \mathcal{L}'
          \begin{cases}
            A & \textit{true} \\
            G &(\overline{l_{5}} \mathbin{\mathsf{U}} l_{3}) 
            \land~~~ \mathsf{G} (l_{5} \rightarrow \mathsf{X} (\overline{l_{5}} \mathbin{\mathsf{U}} l_{3})) 
            \land~~~ \mathsf{G} (l_{3} \rightarrow \mathsf{X} (\overline{l_{3}} \mathbin{\mathsf{U}} l_{5}))
          \end{cases}
\end{align}

In contrast with (\ref{spec:c1}), the strict order among locations, i.e., $l_3 \rightarrow l_5$, found in (\ref{spec:lprime}) does not allows locations $l_3$ or $l_5$ to be visited more than one time per each round of visits.
We have that $\mathcal{L}' \preceq \mathcal{Q}$. However neither $\hat{\mathcal{L}}$ nor $\mathcal{L}'$ refine $\mathcal{C}_1$. Note that (\ref{spec:lprime}) only impose a strict visit order, but does not impose to actually visit the locations. To complete the refinement process we produce a new contract $\mathcal{L}^* = \hat{\mathcal{L}} \parallel \mathcal{L}'$ which can now refine $\mathcal{C}_1$. 



\end{example}

\subsection{Specification Repair via Separation and Merging}
Instead of finding this missing element from $\Delta$, the \emph{repair} operation attempts to make a minimal modification of the top-level specification $\mathcal{C}$ so that the library $\Delta$ can implement it. We use the contract operations of separation and merging for this purpose.

Let $\hat{\mathcal{L}}$ be the candidate composition of library $\Delta$, as before. 
As shown in Figure~\ref{fig:flowdiagram}, we first compute the \textit{separation} between $\hat{\mathcal{L}}$ and $\mathcal{C}$, i.e., $\mathcal{S} = \hat{\mathcal{L}} \div \mathcal{C}$. Note that the order of the element in the dividend ($\hat{\mathcal{L}}$) and the divisor ($\mathcal{C}$) is \textit{opposite} with respect to the order that they had in the quotient. Once we have computed $\mathcal{S}$, we \textit{merge} it with $\mathcal{C}$, generating a new contract $\mathcal{C}'$. Now we can repair $\mathcal{C}$ by \textit{replacing} it with $\mathcal{C}'$, which is guaranteed to refine the candidate composition $\hat{\mathcal{L}}$. 

In contrast to the quotient, the operation of separation, combined with merging, will give us the \textit{smallest abstraction} of $\hat{\mathcal{L}}$ such that $\mathcal{C}$ merged with $\mathcal{S}$ can refine it (see \cite{passeronemerging}).



\begin{example}

   Let us consider the contract $\mathcal{C}_2$, another simplified version of $\mathcal{C}$, which only prescribes that the robot always greets \textit{immediately} when a person is detected (i.e., in the same time-step), assuming that there will always eventually be people detected. We have the following LTL contract:
    
   \begin{equation}
     \mathcal{C}_{2}
         \begin{cases}
           A & \mathsf{GF}(s)\\
           G & \mathsf{GF}(s) \rightarrow \mathsf{G}(s \rightarrow g)
         \end{cases}
         \notag
   \end{equation}
   
   Let us assume that in our library the candidate composition $\hat{\mathcal{L}}$ is the following contract:
   
   \begin{equation}
     \hat{\mathcal{L}}
         \begin{cases}
            A & true \\
            G & \mathsf{G}(s \rightarrow \mathsf{X} g)
         \end{cases}
         \notag
   \end{equation}
   
   $\hat{\mathcal{L}}$ requires  the robot to greet the person in the \textit{next} time instant of when a person is detected. Obviously, $\hat{\mathcal{L}}$ fails to refine $\mathcal{C}_{2}$. Let us now compute the separation between $\hat{\mathcal{L}}$ and $\mathcal{C}_{2}$, obtaining the following contract:

   \begin{equation}
     \mathcal{S}
         \begin{cases}
            A & \mathsf{G} (\mathit{s} \rightarrow \mathit{g}) \lor  \overline{(\mathsf{G} (\mathit{s} \rightarrow \mathsf{X} \mathit{g}) \land \mathsf{G} \mathsf{F} \mathit{s})} \\
            G & \mathsf{G} (\mathit{s} \rightarrow \mathsf{X} \mathit{g}) \land \mathsf{G} \mathsf{F} \mathit{s}
         \end{cases}
         \notag
   \end{equation}
   
    The result of $\mathcal{S}$ merged with $\mathcal{C}_{2}$ is the following contract:
   
   \begin{equation}
     \mathcal{C}_{2}'
         \begin{cases}
            A & \mathsf{G} \mathsf{F} \mathit{s} \land (\mathsf{G} (\mathit{s} \rightarrow \mathit{g}) \lor  \overline{(\mathsf{G} \mathsf{F} \mathit{s} \land \mathsf{G} (\mathit{s} \rightarrow \mathsf{X} \mathit{g}))}) \\
            G & \mathsf{GF}(s) \rightarrow \mathsf{G}(s \rightarrow \mathsf{X}g)
         \end{cases}
         \notag
   \end{equation}


We have `patched' $\mathcal{C}_2$ by creating a new contract $\mathcal{C}_2'$ that can substitute it.
The contract $\mathcal{C}_2'$ can now require the robot to greet on the step after it sees a person, under the assumptions of $\mathcal{C}_2$. Note that the process of generating $\mathcal{C}_2'$ has been fully automatic. It did not require the designer to look at the specifications and make manual adjustments, which can be hard to do as the complexity of the specifications increases. 



\end{example}

\section{Discussion}
\label{sec:discussion}
This section discusses trade-offs that the designer might consider when using \NAME. We consider (i) whether or not the time-steps between the abstract and concrete maps should be the same and (ii) whether the choice of candidate composition should be based on the highest or lowest refinement scores.

\paragraph{Time Step Duration}
Let us consider the specification $\mathsf{OrderedPatrolling}(l_b, l_f, l_e)$,  which requires the robot to patrol in the abstract map locations $L_B, L_F, L_E$ in the order ($L_B \rightarrow L_F \rightarrow L_E$). Specifically, the robot must move away from $L_B$, $L_F$, or $L_E$ immediately after (in the next step) they have been visited and cannot return to the same location before having finished the patrolling of all three. This specification is consistent in the abstract domain. However, it can not be `refined' by any library of components defined over the concrete map. This is the problem:
after visiting $L_3$, in the next time step the robot should leave $L_F$ without going back to $L_5$. However since $L_F$ is \textit{covered} by $L_1$, $L_3$ and $L_4$, the robot is stuck and can never reach $L_2$ (i.e., $L_E$ in the abstract map), thus failing to realize the specification. 

\NAME can help the designer identifies such problems and automatically repair the specification. For example, a candidate composition that prescribes the patrolling of locations $L_5 \rightarrow L_3 \rightarrow L_4 \rightarrow L_2$ can be used to \textit{repair} the abstract specification. This would result in a more \textit{relaxed} $\mathsf{Patrolling}$ of locations, i.e., one that does not require a strict order.
However, a designer might want to consider that a time step in the abstract map has `a different duration' from a time step in the concrete map. Instead of letting \NAME relax the specification by removing the order among locations, the designer could manually repair the specification by, for example, substituting each `\textit{next}' ($\mathsf{X}$) operator with as many next operators as the number of locations in the concrete map. This would ensure that the robot has time to leave the front area in the more concrete description of the store.

\paragraph{Refinement Score}
When looking for implementations in our library that meet the top-level contract, our framework prefers the most refined implementation possible, i.e., this is the implementation supporting as many features as possible. One could argue that this would likely be the most expensive implementation and that, thus, one would prefer the least feature-rich implementation. Our framework can be extended to support this implementation, too. 

When the library cannot immediately refine the top-level specification, we argue that the choice of which candidate composition to choose (i.e., the one with the highest or the lowest refinement score) comes at a trade-off with the strategy adopted (i.e., search or repair). If the strategy is to search for missing components, one would like to have a candidate with the highest refinement score, as this would be a solution that delegates as little functionality as possible to the missing specification that needs to be implemented with an external library. On the other hand, if the strategy is to repair, one could select the composition with the lowest refinement score. By choosing the composition with the least functionality, the `\textit{patch}' that we are applying to the original contract (after performing contract separation and merging) will be `lighter' (i.e., be less demanding) than a repair performed by a more refined candidate composition.

\paragraph{Conclusions}
We presented a contract-based framework for modeling and refining robotic mission specifications using libraries of mission components at various abstraction layers. When the refinement of a specification is not possible out of the current library, we provided a method to automatically repair the specification, so that it can be refined using the library, or effectively guide the search for new implementations that can refine it. Our methodology is fully automated and based on contract manipulations via the quotient, separation, and merging operations. We implemented our framework in the tool \NAME. As future work, we plan to test it on a large-scale case study and further investigate the systematic generation of libraries for robotic mission specification.

\bibliographystyle{splncs04}
\bibliography{references}


\appendix
\section{Generating the context constraints}\label{sec:appendix}

In our modeling framework, types are used to generate world context constraints semantically.
For each type of relationship described above, our framework produces an LTL formula which can be added to the world context.

In the following discussion, for a formula $\varphi$, we will use $AP^\varphi$ to denote the set of atomic propositions that appear in the syntax of $\varphi$; we will call the \emph{types of $\varphi$} the set of types associated with the atomic propositions that appear in $\varphi$. For example, if $\varphi = l_1$, $AP^\varphi = \{l_1\}$, and the types of $\varphi$ is the set $\{L_1\}$.
We define the following functions:
\begin{itemize}
 \item $MTX(\varphi)$ produces an LTL formula enforcing the mutual exclusivity conditions of the types of $\varphi$. For any atomic propositions $p_i, p_j \in AP^{\varphi}$, where $P_i$ and $P_j$ are mutually exclusive types, we append the constraint $\mathsf{G}(p_i \rightarrow \overline{p_j}) \land \mathsf{G}(p_j \rightarrow \overline{p_i})$, i.e.,
 \begin{equation}
  MTX(\varphi) = \bigwedge_{\substack{p_i, p_j \in AP^{\varphi}\\P_i \textit{mutex with} P_j}} \mathsf{G}(p_i \rightarrow \overline{p_j}) \land \mathsf{G}(p_j \rightarrow \overline{p_i}).\notag
 \end{equation}
 \item $ADJ$ produces an LTL formula enforcing the adjacency conditions of all adjacent types. For any atomic propositions $p_i, p_j$, where $P_i$ and $P_j$ are related by an adjacency relation, we include the constraint $\mathsf{G}(p_i \rightarrow \mathsf{X} (p_i \lor p_j)) \land \mathsf{G}(p_j \rightarrow \mathsf{X} (p_j \lor p_i))$, i.e.,
  \begin{equation}
  ADJ = \bigwedge_{\substack{P_i \text{ adj. to } P_j}} \left(\begin{aligned}& \mathsf{G}(p_i \rightarrow \mathsf{X} (p_i \lor p_j)) \land \\ & \mathsf{G}(p_j \rightarrow \mathsf{X} (p_j \lor p_i))\end{aligned}\right).
 \end{equation}
 \item $EXT$ produces an LTL formula enforcing all extension relations. That is, for any atomic propositions $p_i, p_j$, where $P_i \preceq P_j$, $EXT(\varphi)$ includes the clause $\mathsf{G}(p_i \rightarrow p_j)$, i.e.,  
 \begin{equation}
  EXT = \bigwedge_{\substack{P_i \preceq P_j}} \mathsf{G}(p_i \rightarrow p_j).
 \end{equation}
\item $COV$ produces an LTL formula enforcing the coverage constraints among all types with such a constrained defined. In other words, for any atomic propositions $p_a, p_b$, where $b \in \mathcal{I}$ (an indexing set) such that $\{P_b\}_{b \in \mathcal{I}}$ covers $P_a$, we include the constraint $\mathsf{G}( p_a \to \bigvee_{b \in \mathcal{I}} p_b )$, that is,
\begin{equation}
COV = \bigwedge_{\substack{\{P_b\}_{b \in \mathcal{I}} cov. P_a}} \mathsf{G}\left( p_a \to \bigvee_{b \in \mathcal{I}} p_b \right).
\end{equation}
\end{itemize}

\begin{example}
In our running example, we have two maps at two levels of abstraction. There is an `abstract map' with locations $L_B$, $L_F$, and $L_E$. And there is a `concrete map' with locations $L_1$, $L_2$, $L_3$, $L_4$ and $L_5$.

We assign a type to every location on the map. This type is instantiated as an atomic proposition, e.g., $L_5$ has an associated atomic proposition $l_5$. Whenever $l_5$ is \textit{true}, the robot is in location $L_5$ on the map.
We also define the type $S$ to model a sensor that detects the presence of a person and the type $G$ to model the greeting action. We use $s$ and $g$ for the atomic propositions corresponding to types $S$ and $G$, respectively.

For every type we define its mutual exclusion, adjacency, extension, and covering relationships. For example, $L_1$ has a \textit{adjacency} relationship with types $L_2$ and $L_3$; it is mutually exclusive with $L_2, L_3, L_4$, and $L_5$ since the robot cannot be in multiple locations at the same time; $L_1$ \textit{extends} $L_F$, i.e., $L_1 \preceq L_F$; and $L_1$ is part of the $\{L_1, L_3, L_4\}$ covering of $L_F$.

\end{example}

\subsection{Verifying specifications}
Once the designer uses \NAME to define the types as discussed above, many relationships between atomic propositions are automatically inferred, according to the expressions for $MXT$, $ADJ$, $EXT$, and $COV$. \NAME can perform three types of checks: consistency, refinement, and realizability.

\NAME performs consistency checks on the mission specification and all the components in the library. Consistency means that formulas are satisfiable. 
Refinement checks are performed to verify whether a specification is more stringent than another. This is particularly important when checking whether a composition of elements from the library can meet a top-level specification. Reliazability means that a specification can be implemented such that it behaves according to the specification for all possible inputs of its uncontrolled variables.

\paragraph{Consistency Check} 
In a consistency check, we only consider the context constraints to be $MTX$ and $ADJ$, since we only want to prove that the formulas of a single contract are satisfiable.
For any contract having $\varphi_A$ and $\varphi_G$ as assumptions and guarantees, we check that both $\varphi_A$ and $\varphi_G$ are consistent by proving the satisfiability of the following formulas:
\begin{align*}
  \varphi_A & \land MTX(\varphi_A) \land ADJ(\varphi_A)\\
  \varphi_G & \land MTX(\varphi_G) \land ADJ(\varphi_G)
\end{align*}

For example, let $l_b$ and $l_f$ be the atomic propositions of locations $L_B$ and $L_F$. If the designer formulates a specification having as guarantees $l_b \land l_f$ and \textit{true} assumptions, \NAME checks that $l_b \land l_f \land \mathsf{G}(l_b \rightarrow \overline{l_f}) \land \mathsf{G}(l_f \rightarrow \overline{l_b}) \land \mathsf{G}(l_b \rightarrow \mathsf{X}(l_b \lor l_f)) \land \mathsf{G}(l_f \rightarrow \mathsf{X}(l_f \lor l_b))$ has no satisfiable assignments, proving that the contract is inconsistent. \ins{Note that we have not included the adjacency relationships of all types for simplicity.}

\paragraph{Refinement Check} For the refinement verification, we need to take in consideration the context constraints given by $EXT$ and $COV$ because these connect abstract types with their concrete subtypes and coverings. Let $\mathcal{C}_1=(\varphi_{A1}, \varphi_{G1})$ and $\mathcal{C}_2=(\varphi_{A2}, \varphi_{G2})$ be two contracts. In order to prove that $\mathcal{C}_1 \preceq \mathcal{C}_2$, we have to check whether $\varphi_{G1} \rightarrow \varphi_{G2}$ and $\varphi_{A2} \rightarrow \varphi_{A1}$ are \textit{valid} formulas (we assume the guarantees to always be in their saturated form). 


\NAME, for any \textit{validity check} of a formula, $\phi = \varphi_1 \rightarrow \varphi_2$ first checks the \textit{satisfiability} of the formulas:
\begin{align}
  \varphi_1 & \land MTX(\varphi_1) \land ADJ(\varphi_1) \label{eq:sat1}\\
  \varphi_2 & \land MTX(\varphi_2) \land ADJ(\varphi_2) \label{eq:sat2}
\end{align}

If they are satisfiable, we then proceed to verify the \textit{validity} of the implication $\phi$ in the world context:

\begin{equation}
  EXT \land COV \rightarrow (\phi) \label{eq:extcov}
\end{equation}



\begin{example}
Suppose that we want to check whether the robot by $\mathsf{Patrolling}$ locations $L_1$ and $L_3$ in the concrete map is also $\mathsf{Patrolling}$ location $L_F$ in the abstract map. $\mathsf{Patrolling}$ is a robotic pattern~\cite{menghipatterns} that requires the robot to visit locations infinitely often. That is, we want to prove that $(\mathsf{G}\mathsf{F}(l_1) \land \mathsf{G}\mathsf{F}(l_3))$ is a refinement of $\mathsf{G}\mathsf{F}(l_f)$.

\NAME, after checking the satisfiability of the formulas \eqref{eq:sat1} and \eqref{eq:sat2}, proves the validity of the formula

{\small
\begin{equation}
  (\mathsf{G}(l_1 \rightarrow l_f) \land \mathsf{G}(l_3 \rightarrow l_f)) \rightarrow ((\mathsf{G}\mathsf{F}(l_1) \land \mathsf{G}\mathsf{F}(l_3)) \rightarrow \mathsf{G}\mathsf{F}(l_f)). \label{eq:exrefcheck}
\end{equation}}

Since \eqref{eq:exrefcheck} is valid, we can conclude that a robot, by \textit{visiting} the locations $L_1$ and $L_3$ infinitely often, is also visiting location $L_F$ infinitely often, connecting a concrete specification to a more abstract one.

\end{example}

\begin{remark}
Note that the formula in \eqref{eq:exrefcheck} is a simplified version the formula in \eqref{eq:extcov}. We do not always need the context to contain constraints enforcing coverage and extensions among \textit{all} types. In this example, it is sufficient to have the context containing the extension relationships among $l_1$, $l_3$ and $l_f$ to prove the refinement. However, if the formula $\phi$ in \eqref{eq:extcov} is \textit{not} monotonic, meaning that some of the atomic propositions appear negated and some not, then it is necessary to add $COV$ to the context constraints.
\end{remark}

\paragraph{Realizability Check} We say that a contract $\mathcal{C} = (\varphi_A, \varphi_G)$ is \textit{realizable} if the formula
\begin{equation*}
  \phi = MTX(\varphi_A) \land ADJ(\varphi_A) \rightarrow  MTX(\varphi_G) \land ADJ(\varphi_G)
\end{equation*}
 can produce a finite state machine that implements it via reactive synthesis. The context constraints of $EXT$ and $COV$ are not needed because the contract to realize is on a unique `\textit{abstraction level}'. For example, to implement the $\mathsf{Patrolling}$ of locations $l_1$ and $l_3$ in the previous example, \NAME checks the realizability of the following formula:
\begin{align}
  &\mathsf{G}(l_1 \rightarrow \overline{l_3}) \land \mathsf{G}(l_3 \rightarrow \overline{l_1}) \land \notag\\
  &
  \land \mathsf{G}(l_1 \rightarrow \mathsf{X}(l_1 \lor l_2 \lor l_3)) \land \mathsf{G}(l_3 \rightarrow \mathsf{X}(l_3 \lor l_1 \lor l_4 \lor l_5)) \land \notag\\
  &
  \land \mathsf{G}\mathsf{F}(l_1) \land \mathsf{G}\mathsf{F}(l_3) \notag
\end{align}

Our framework automatically checks the consistency of every contract and all the refinement relationships among them. These checks are translated into model checking problems, and NuSMV~\cite{nuxmv} is used to solve them. We use STRIX~\cite{strix} to check the realizability of contracts \ins{in the library (if their implementation is missing)} and to produce Mealy machines that implement them when they are realizable.\pierg{Add comment on when it's not realizable we search and repair}

\end{document}